\documentclass[letterpaper, 10 pt, journal, twoside]{ieeetran}
\usepackage{amsmath,amsfonts}
\usepackage{algorithmic}
\usepackage{algorithm}
\usepackage{array}
\usepackage[caption=false,font=normalsize,labelfont=sf,textfont=sf]{subfig}
\usepackage{textcomp}
\usepackage{stfloats}
\usepackage{url}
\usepackage{verbatim}
\usepackage{graphicx}
\usepackage{cite}

\usepackage[bookmarks=true]{hyperref}
\usepackage{hyperref}
\usepackage{url}            %
\usepackage{cleveref}
\usepackage{array, makecell, multirow, booktabs}
\usepackage{times}
\usepackage{framed}
\usepackage[normalem]{ulem}
\usepackage{pdfpages}

\usepackage{amsthm}
\usepackage{amsfonts,bm}       %
\usepackage{amssymb}
\usepackage{siunitx}
\usepackage{amsmath}
\usepackage{bbm}

\usepackage{enumitem}
\usepackage{newtxmath}
\usepackage{float}
\usepackage{changepage}
\usepackage{mathtools}
\graphicspath{ {images/} }
\usepackage[normalem]{ulem}

\renewcommand\fbox{\fcolorbox{red}{white}}
\definecolor{difcolor}{RGB}{0,0,0}
\definecolor{difcolor_border}{RGB}{255,255,255}

\newcommand{\va}{{\bf a}}

\newcommand{\vp}{{\bf p}}
\newcommand{\vs}{{\bf s}}
\newcommand{\vq}{{\bf q}}

\newcommand{\vx}{{\bf x}}

\newcommand{\vtheta}{\boldsymbol{\theta}}

\definecolor{mygreen}{RGB}{0,204,102}
\definecolor{mycolor}{RGB}{211,118,118}
\definecolor{myblue}{RGB}{0,102,204}

\newcommand\mydots{\hbox to 1em{.\hss.\hss.}}
\newcommand{\ind}{\mbox{\(\mathbf{1}\)}}

\begin{document}

% \usepackage{titlesec}
% \raggedbottom
%%%%%%%%%%%%%%%%%%%%%%%%%%%%%%%%%%%%%%%%%%%%%%%%%%%%%%%%%%%%%%%%%%%%%%%%%%%%%%%%
\title{ILCL: Inverse Logic-Constraint Learning from\\
Temporally Constrained Demonstrations}

%%%%%%%%%%%%%%%%%%%%%%%%%%%%%%%%%%%%%%%%%%%%%%%%%%%%%%%%%%%%%%%%%%%%%%%%%%%%%%%%
\author{
Minwoo Cho\textsuperscript{1}, 
Jaehwi Jang\textsuperscript{1,2}, 
and Daehyung Park\textsuperscript{1\textdagger}\vspace{-30pt}%
\thanks{Manuscript received June 11, 2025; Revised September 26, 2021.; Accepted October 24, 2025. This paper was recommended for publication by Editor Jens Kober upon evaluation of the Associate Editor and Reviewers’ comments.
This work was partly supported by Institute of Information \& communications Technology Planning \& Evaluation (IITP) grant funded by the Korea government (MSIT) (No. RS-2022-II220311, RS-2024-00509279, and RS-2024-00336738), the Technology Innovation Program funded by the Ministry of Trade, Industry \& Energy (MOTIE) (RS-2024-00423940), and the KAIST Convergence Research Institute Operation Program.
}
\thanks{
\textsuperscript{1}The authors are with Korea Advanced Institute of Science and Technology, Korea ({\tt\small \{cmw9903, daehyung\}@kaist.ac.kr}). \textsuperscript{\textdagger}D. Park is the corresponding author. \textsuperscript{2}The author is with Georgia Institute of Technology, USA {\tt\small \{ jjang318@gatech.edu\}}.}
\thanks{Digital Object Identifier (DOI) : see top of this page.}
}
\markboth{IEEE ROBOTICS AND AUTOMATION LETTERS. PREPRINT VERSION. ACCEPTED OCTOBER, 2025}%
{Cho \MakeLowercase{\textit{et al.}}: Inverse Logic-Constraint Learning from
Temporally Constrained Demonstrations}

%%%%%%%%%%%%%%%%%%%%%%%%%%%%%%%%%%%%%%%%%%%%%%%%%%%%%%%%%%%%%%%%%%%%%%%%%%%%%%%%

\maketitle
% \thispagestyle{empty} 
% \pagestyle{empty}
%%%%%%%%%%%%%%%%%%%%%%%%%%%%%%%%%%%%%%%%%%%%%%%%%%%%%%%%%%%%%%%%%%%%%%%%%%%%%%%%
\begin{abstract}
We aim to solve the problem of temporal-constraint learning from demonstrations to reproduce demonstration-like logic-constrained behaviors. Learning logic constraints is challenging due to the combinatorially large space of possible specifications and the ill-posed nature of non-Markovian constraints. To this end, we introduce inverse logic-constraint learning (ILCL), a novel temporal-constraint learning method formulated as a two-player zero-sum game between 1) a genetic algorithm-based temporal-logic mining (GA-TL-Mining) and 2) logic-constrained reinforcement learning (Logic-CRL). GA-TL-Mining efficiently constructs syntax trees for parameterized truncated linear temporal logic (TLTL) without predefined templates. Subsequently, Logic-CRL finds a policy that maximizes task rewards under the constructed TLTL constraints via a novel constraint redistribution scheme. Our evaluations show ILCL outperforms state-of-the-art baselines in learning and transferring TL constraints on four temporally constrained tasks. We also demonstrate successful transfer to real-world peg-in-shallow-hole tasks.
\end{abstract}


% %%%%%%%%%%%%%%%%%%%%%%%%%%%%%%%%%%%%%%%%%%%%%%%%%%%%%%%%%%%%%%%%%%%%%%%%%%%%%%%%
\section{Introduction}
\begin{figure}[t]    
\centering
\includegraphics[width=0.95\columnwidth]{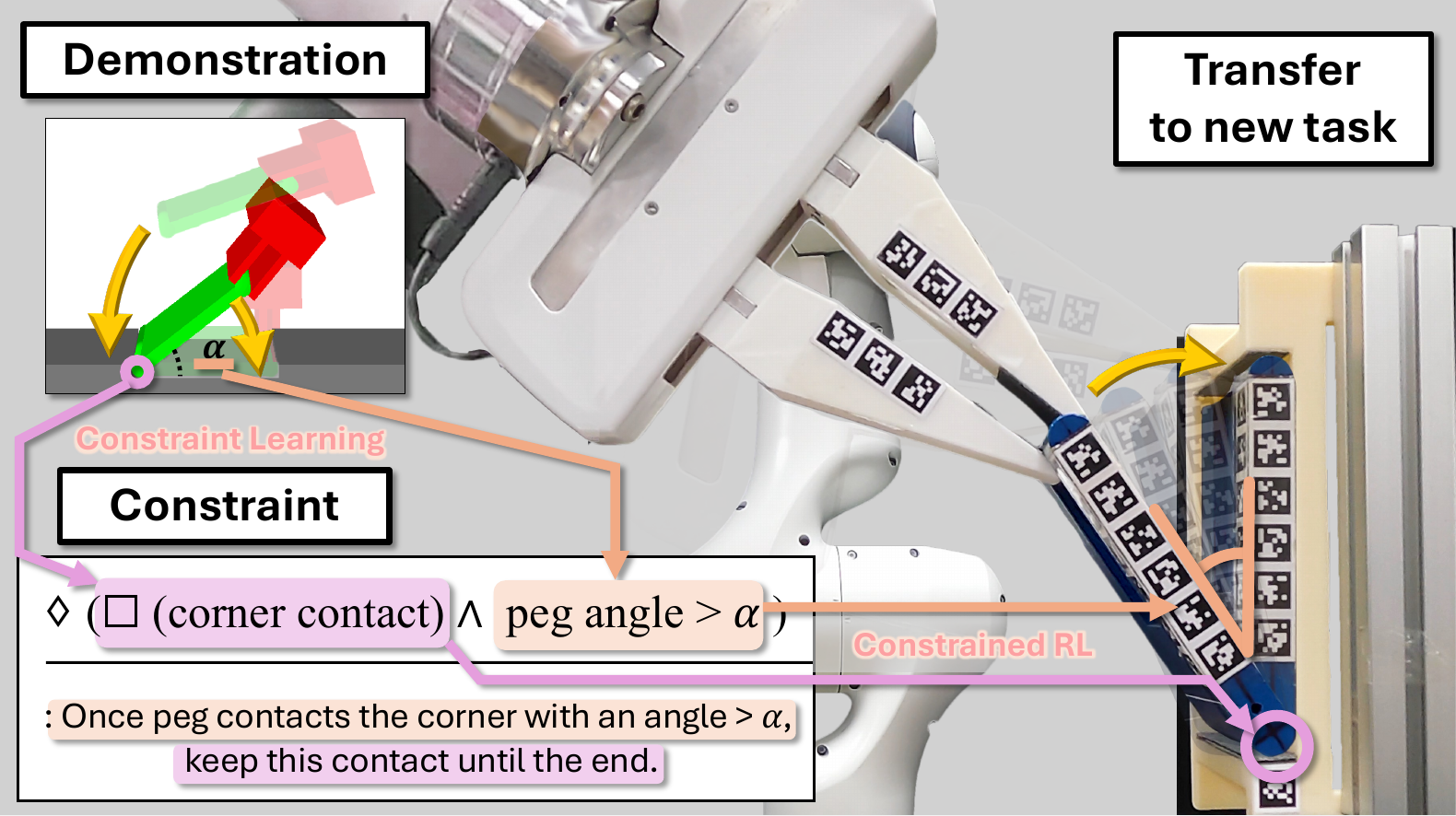}
    \caption{
    An exemplar \textit{peg-in-shallow-hole} task, where one corner of the peg must maintain contact with the hole during insertion using a parallel jaw gripper. By learning temporal logic constraints from demonstrations, our method successfully transfers the constrained insertion behavior to an unseen, tilted hole environment.
    }
    \label{fig_main} 
    \vspace{-10pt}
\end{figure}

\IEEEPARstart{R}{obot} policy learning often involves constraints that account for user, task, and environmental restrictions, as shown in Fig.~\ref{fig_main}. To autonomously adopt constraints, inverse constraint learning (ICL) methods recover constraints from demonstration~\cite{liu2025a}. While ICL typically targets global constraints, most tasks require spatial or temporal restrictions. For example, mobile robots must wait until the traffic sign turns green. To express such structured and interpretable restrictions, researchers often adopt temporal logic (TL), such as linear temporal logic (LTL) \cite{ltlpnueli1977temporal} and signal temporal logic (STL) \cite{stlmaler2004monitoring}.

In this context, we aim to solve the problem of TL-constraint learning from demonstrations to reproduce demonstration-like constraint-satisfying behaviors. However, TL-constraint learning is challenging due to the combinatorially large space of logic specifications in addition to the non-differentiable representation following strict syntactic rules. For efficient searches, researchers often rely on predefined logic templates~\cite{yifru2024concurrent} or restrict operator sets~\cite{liu2024interpretable}. However, the limited expressivity restricts applications in tasks. Therefore, a desired approach requires to derive free-form logic constraints. 

Another challenge arises from the ill-posed nature of non-Markovian constraints. Given this ill-posedness, conventional approaches often focus on finding constraints without simultaneously considering the reward maximization in policy learning. This reduces the chance of finding optimal and transferable constraints from experts. Further, the non-Markovian property requires evaluating the history of state-action sequences, which yields sparse feedback during policy learning. These combined challenges complicate the integration of TL with conventional ICL frameworks~\cite{liu2025a} via inverse and forward constraint reinforcement learning (CRL) processes.

% solution
We propose inverse logic-constraint learning (ILCL), a free-form transferable TL-constraint learning algorithm. Our method involves constraint learning as a two-player zero-sum game~\cite{kim2024learning}, between 1) a genetic algorithm-based temporal-logic mining (GA-TL-Mining) and 2) logic-constrained reinforcement learning (Logic-CRL). GA-TL-Mining constructs a syntax tree for parameterized truncated linear temporal logic (TLTL)~\cite{tltlli2017reinforcement} without predefined templates. Subsequently, Logic-CRL finds a policy that maximizes task rewards while adhering to the constructed TLTL constraints, measuring the degree of logic satisfaction across a trajectory using a robustness function. To handle the history dependency inherent in TL constraints in CRL, we introduce a constraint redistribution method, analogous to reward redistribution approaches \cite{arjona2019rudder,gangwani2020learning}, converting trajectory constraints into state-action constraints. To the best of our knowledge, we provide the first CRL approach that generalizes to arbitrary TL constraints.

We evaluate ILCL against state-of-the-art ICL baselines in four simulated navigation and manipulation benchmark environments. ILCL outperforms baselines in learning diverse forms of TL constraints, achieving the lowest constraint-violation scores while yielding high task reward returns comparable to expert behaviors. We also demonstrate the applicability of the learned constraints by successfully transferring a constraint to a real-world \textit{peg-in-shallow-hole} task, thereby demonstrating the robustness and scalability of TL constraints.

\section{Related Work}
\noindent\textbf{Constraint learning and mining.} ICL recovers constraints from demonstrations. Early work learns numerical constraints in discretized states~\cite{scobeemaximum}, while, with neural representations, recent methods model constraints on continuous state or state-action spaces~\cite{malik2021inverse, jang2023inverse}. To scale, researchers introduce multi-task~\cite{kim2024learning} and multi-modal~\cite{qiao2024multi} approaches. However, numerical representations are often hard to interpret and transfer. Instead, we use logical representations for robust generalization. 

Logic constraint learning is a variant of ICL. Early methods identify logic formulas adopting inductive logic programming~\cite{baert2023learning}, one-class decision tree~\cite{baert2024learning}, or data-driven rule induction~\cite{kusters2022differentiable}. However, these approaches yield simple global propositions (e.g., ${}^{\forall}(x<3)$) or timestep-specific constraints, thereby restricting generalizability. To enhance temporal representation, researchers introduce TL in imitation learning, but algorithms still face the same limitation of relying on predefined logic forms~\cite{yifru2024concurrent}, \cite{wang2023temporal} or limited temporal expressions~\cite{liu2024interpretable}. In contrast, our method performs specification mining without relying on pre-defined templates.

Logic specification mining infers candidate system properties from observations (e.g., STL mining~\cite{bartocci2022survey}). To handle the combinatorially large specification space, researchers often fix logic structures~\cite{jha2017telex}, discretize the search space~\cite{vaidyanathan2017grid}, or restrict the search depth~\cite{mohammadinejad2020interpretable}. Alternatively, researchers enable tree-based genetic algorithms to explore the space of logic syntax trees without restrictions~\cite{nenzi2018robust}. We extend the algorithm to compose a TL constraint with additional temporal operators. 

\noindent\textbf{Constrained Reinforcement Learning.} CRL, also known as safety RL, aims to find a reward-maximizing policy while satisfying safety constraints. Garcia and Fernandez classify approaches into two categories: 1) modification of the exploration process and 2) the modification of the optimization criterion~\cite{garcia2015comprehensive}. The first category typically restricts actions or policies within a trust region~\cite{achiam2017constrained}. Although these connect to formal verification, the temporal constraints we use make it hard to define the trust space for action sequences. 

On the other hand, the second category often employs Lagrangian relaxation to convert the constrained optimization in CRL into a typical RL problem with unconstrained objectives (i.e., rewards)~\cite{tessler2018reward}.
However, the long-term dependency on TL may result in reward delays. To address this, researchers introduce neural networks to estimate trajectory returns~\cite{arjona2019rudder} or redistribute rewards across time steps, smoothing the reward signals~\cite{gangwani2020learning}. We extend this redistribution to TL-based CRL, evaluating constraint violations at the end of episodes.

In addition, TL-based CRL approaches often prioritize logic satisfaction, neglecting the primary task objective within environments. To mitigate this, approaches weigh logic-based rewards~\cite{Shah2024ltlconstrained} or apply logic fragments only~\cite{Wen2015CorrectbysynthesisRL}. These strategies limit their generalizability across novel environments and the adoption of complex TL specifications. In contrast, our method operates on arbitrary TL in a model-free manner.

\section{Preliminaries}
%%%%%%%%%%%%%%%%%%%%%%%%%%%%%%%%%%%%%%%%%%%%%%%%%%%%%%%%%%%%
\vspace{-2pt}
\subsection{Constraint Markov Decision Process (CMDP)}
\vspace{-2pt}
A CMDP is a tuple $(\mathcal{S}, \mathcal{A}, P,  R, C)$, where $\mathcal{S}$ is the state space, $\mathcal{A}$ is the action space, $P: \mathcal{S} \times \mathcal{A} \times \mathcal{S} \rightarrow \mathbb{R}^+$ is the transition function, $R: \mathcal{S} \times \mathcal{A} \rightarrow \mathbb{R}$ is the reward function, and $C: \mathcal{S}\times \mathcal{A} \rightarrow \mathbb{R}$ is the constraint-cost function. 
A policy $\pi: \mathcal{S} \times \mathcal{A} \rightarrow \mathbb{R}^+$ defines a probability density function to determine the likelihood of selecting an action given a state. 
Then, the objective of CMDP is to find a policy $\pi^*$ that generates a state-action sequence $\xi=(\vs_0, \va_0, ..., \vs_T, \va_T)$ with a finite horizon $T$ maximizing the expected return while the expectation of cumulative constraint cost bounded to a budget $d$: $\pi^*=\arg\max_\pi \mathbb{E}_{\xi\sim\pi}[\sum^{T}_{t=0}R(\vs_t, \va_t)]$ subject to $\mathbb{E}_{\xi\sim\pi}[\sum^{T}_{t=0} C(\vs_t, \va_t)]\leq d$. In this work, we use $R(\xi)=\sum^{T}_{t=0} R(\vs_t, \va_t)$ and $C(\xi)=\sum^{T}_{t=0} C(\vs_t, \va_t)$ to denote the cumulative rewards and constraint costs, respectively.

To generate a sequence satisfying TL constraints, we introduce another CMDP with a TLTL constraint $C_\phi$, which returns a positive value when a state sequence $\xi^{(s)}=(\vs_0, ..., \vs_T)$ violates the TLTL formula $\phi$ at any time. For brevity, we use state-action sequence, state sequence, and trajectory interchangeably and omit the arguments of $C(\vs,\va)$ and $C(\xi)$.

\begin{figure*}[t]
\centerline{\includegraphics[width=0.95\textwidth]{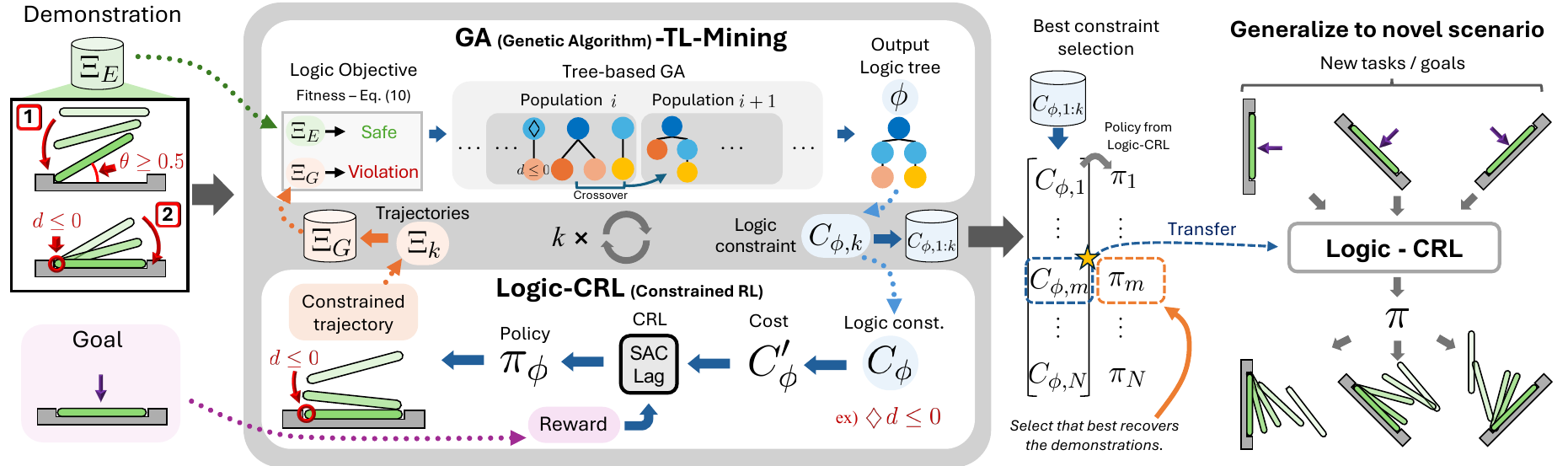}}
    \caption{
    Illustration of ILCL that learns TLTL constraint from demonstration to generalize to novel scenario. ILCL consists of two parts: \textbf{GA-TL-Mining} (Middle top) which generates a TLTL constraint that distinguishes demonstration trajectories from generated trajectories through a genetic algorithm on the TL syntax tree; and \textbf{Logic-CRL}, which optimizes the policy for task rewards under the generated TL constraint using SAC-Lag~\cite{ha2021learning}. Once ILCL identifies a TL constraint, Logic-CRL derives a new policy applicable to novel scenarios.
    }
    \label{fig:overall} 
    \vspace{-1.5em}
\end{figure*}

%%%%%%%%%%%%%%%%%%%%%%%%%%%%%%%%%%%%%%%%%%%%%%%%%%%%%%%%%%%%
\vspace{-5pt}
\subsection{Truncated Linear Temporal Logic (TLTL)}
\vspace{-1pt}
We employ TLTL~\cite{tltlli2017reinforcement}, a predicate temporal logic that extends LTL~\cite{ltlpnueli1977temporal} and LTL over finite traces ($\text{LTL}_f$~\cite{de2013linear}). Each TLTL formula consists of atomic predicates (AP) of the form $a\cdot \mathbf{s}^{(i)}< b$, where $a\in\{-1, 1\}$, $b\in\mathbb{R}$ is a constant, and $\mathbf{s}^{(i)}$ refers to the $i$-th element of the state vector $\vs$. The syntax of TLTL formulas with an AP $\mu$ is defined as follows: 
\begin{align}
\label{eq:pnf}
    \phi ::= \top \,|\, \bot &\,|\, \mu \,|\, \neg \mu \,|\, \phi_1 \land \phi_2 \,|\, \phi_1 \lor \phi_2 \,|\, \bigcirc \phi \,|\, \nonumber \\ &\lozenge \phi \,|\, \square \phi \,|\, \phi_1 \mathcal{U} \phi_2 \,|\, \phi_1 \mathcal{R} \phi_2,
\end{align}
where $\top$ and $\bot$ denote Boolean True and False; $\neg$ (negation), $\land$ (conjunction), and $\lor$ (disjunction) are Boolean connectives; and $\lozenge$ (eventually), $\square$ (always), $\mathcal{U}$ (until), $\mathcal{R}$ (release), and $\bigcirc$ (next) are temporal operators. The syntax in Eq.~\eqref{eq:pnf} follows the positive normal form, where negations apply only to APs. We evaluate TLTL formulas against a finite state sequence $\vs_{t:t+\Delta} \hspace{-1pt}=\hspace{-1pt}(\vs_t, \mydots, \vs_{t+\Delta})$ from time step $t$ with the finite length $\Delta$.

To incorporate learnable thresholds into TLTL, we introduce a parametric variant, we call pTLTL, inspired by parametric STL (pSTL)~\cite{asarin2012parametric}. The parameterization replaces the constant $b$ in each AP with a parameter $\theta$, yielding a parametric AP (pAP) of the form $a \cdot \vs^{(i)}<\theta$. For example, we represent a TLTL formula $\phi$ as a pTLTL formula $\phi_{\vtheta}$:
\begin{align}
\phi &= (\vs^{(0)} < 1) \, \mathcal{U} \, (\vs^{(1)} > 2) && \dots\quad\,\,\, \text{(TLTL)}, \nonumber\\
\phi_{\vtheta} &= (\vs^{(0)} < \vtheta^{(0)}) \, \mathcal{U} \, (\vs^{(1)} > \vtheta^{(1)}) && \dots\quad\text{(pTLTL)},
\end{align}
where a parameter vector $\vtheta = (\vtheta^{(0)}, \vtheta^{(1)}) \in \mathbb{R}^2$. In this work, we use the learnable pTLTL; for simplicity, we use TLTL $\phi$ and pTLTL $\phi_{\vtheta}$ interchangeably and denote them as $\phi$. 

Then, the Boolean semantics of a formula $\phi$ defines a state sequence $\vs_{t:t+\Delta}$ satisfies $\phi$ at time $t$ if and only if $\rho(\vs_{t:t+\Delta}, \phi, t) > 0$, where $\rho$ is the robustness function~\cite{tltlli2017reinforcement}. This function provides quantitative semantics, assigning large positive values for strong satisfaction and large negative values for significant violation. For simplicity, we denote robustness as $\rho(\xi, \phi)$, omitting the time index. 

%%%%%%%%%%%%%%%%%%%%%%%%%%%%%%%%%%%%%%%%%%%%%%%%%%%%%%%%%%%%
\vspace*{-10pt}
\section{Methodology}
\vspace{-1pt}
\subsection{Problem formulation}
The objective of ILCL is to identify a constraint $C_\phi$, represented by a TLTL formula $\phi$, from the expert trajectories $\Xi_E$ that exhibit zero violations. Then, we formulate the identification as a constrained min-max optimization that finds an optimal constraint $C_{\phi}^*$ maximizing the expected cumulative rewards of the inferred constrained policy $\pi$, while minimizing its expectation difference from the expert policy $\pi_E$:
\begin{align}
\label{eq:problem1} 
C_\phi^* =  &\arg\min_{C_\phi} \max_{\pi}
&&\mathbb{E}_{\xi \sim \pi} [R(\xi)] 
   - \mathbb{E}_{\xi \sim \pi_{E}} [R(\xi)]  \\ 
&\quad\text{s.t.} 
&&\mathbb{E}_{\xi \sim \pi} [C_\phi(\xi)] = \mathbb{E}_{\xi \sim \pi_{E}} [C_\phi(\xi)] = 0,  \nonumber 
\end{align}
where ${C_\phi(\xi)\!\!=\!\!\ind [\rho(\xi, \phi)\!\! <\!\! 0]}$. However, this problem is intractable due to the computation of $\pi$ for every candidate $C_\phi$.

Alternatively, we reformulate Eq.~(\ref{eq:problem1}) as a two-player zero-sum game, inspired by inverse state-action constraint learning in~\cite{kim2024learning}. Our key distinction from~\cite{kim2024learning} is the use of a logic-based constraint $C_\phi$ defined over state-action sequences. This formulation alternates between constraint and policy optimization players as shown in Fig.~\ref{fig:overall}. In each $k$-th iteration, the constraint optimization (i.e., GA-TL-Mining) identifies a constraint $C_{\phi,k}$ that maximally penalizes the previously learned policies $\pi_1, \dots, \pi_k$ relative to the expert policy $\pi_E$:
\begin{align}
\label{eq:icl_cl_obj}
 C_{\phi, k}\hspace{-2pt}=&
 \;\underset{C_\phi}{\text{argmax}}
 \hspace{-1pt}
 \underset{i\leq k}{{\Sigma}}
 \mathbb{E}_{\xi \sim \pi_{i}}  [C_\phi(\xi)], \, \textrm{s.t.}\,\,\mathbb{E}_{\xi \sim \pi_{E}} [ C_\phi(\xi)]=0.
\end{align}
The policy optimization (i.e., Logic-CRL) then finds a new optimal policy $\pi_{k+1}$ that satisfies the constraint $C_{\phi,k}$: 
\begin{align}
\pi_{k+1}\hspace{-2pt}=\;&
\underset{\pi}{\text{argmax}}
\; \mathbb{E}_{\xi \sim \pi} \left[ R(\xi) \right], \; \textrm{s.t.} \; \mathbb{E}_{\xi \sim \pi}[C_{\phi,k}(\xi)]=0.
\label{eq:icl_crl_obj}
\end{align}
These alternations progressively find a constraint that separates the expert policy from discovered high reward policies than the expert. This aligns with the ground-truth constraint, as the expert policy obtains the highest reward among policies that satisfy the constraint. As a result, the identified constraint induces a policy that minimally violates the constraint while gaining higher rewards than the expert policy. 
%Below, we describe each player in detail.
\vspace*{-8pt}

\subsection{Constraint player: GA-based temporal-logic mining}
\label{sec:logic_mining}
At iteration $k$ of ILCL, the constraint player\textemdash GA-TL-Mining\textemdash identifies a TLTL constraint $\phi$ that maximizes a fitness score $\mathcal{F}_k(\phi)$ given demonstrations $\Xi_E$. The score $\mathcal{F}_k(\phi)$ is a Monte Carlo approximation of the dual objective in Eq.~(\ref{eq:icl_cl_obj}). This increases when $\phi$ penalizes violations by prior policies $\pi_{1:k}$, while assigning zero if violated by expert trajectories $\Xi_E$;%. We define $\mathcal{F}_k(\phi)$ as
\begin{equation}
\label{eq: GA-fitnessbase}
\mathcal{F}_k(\phi)\!=\!\frac{\sum_{\xi\in\Xi_{1:k}}\ind[\rho(\xi,\phi)\!<\!0]}{|\Xi_{1:k}|} \!\cdot\! \Pi_{\xi_E\in\Xi_E}\ind[\rho(\xi_E, \phi)\!>\!0],
\end{equation}
where $\Xi_{1:k}\!=\!\{\xi |\xi\!\sim\!\pi_i, i\!\in\! [1,k] \}$ and $\pi_i$ denotes policies generated by Logic-CRL. Using the fitness, GA-TL-Mining iteratively improves a population of TLTL constraints, represented as logic trees, via selection, crossover, and mutation, following the tree-based GA framework, ROGE~\cite{nenzi2018robust}. For simplicity, we refer TLTL constraints and logic trees interchangeably.

GA-TL-Mining consists of three steps\textemdash 1) initial population generation, 2) parent selection, and 3) offspring generation:

%\medskip
\noindent\textbf{1) Initial population generation}: 
GA-TL-Mining begins with an initial population of logic trees $\Phi$, which consists of $N_\text{B}$ basis logic trees $\{\phi_i^{\text{B}}\}_{i=1}^{N_\text{B}}$ and $N_\text{R}$ random logic trees $\{\phi_i^{\text{R}}\}_{i=1}^{N_\text{R}}$: 
\begin{align}
\Phi=\{\phi_1^{\text{B}}, \ldots, \phi_{N_\text{B}}^{\text{B}}\} \cup \{\phi_1^{\text{R}}, \ldots, \phi_{N_\text{R}}^{\text{R}}\}.
\end{align}
We construct the basis logic trees over the state space $\mathcal{S} \subseteq \mathbb{R}^{\kappa}$ where $\kappa$ is the dimensionality. For each dimension $i\in[1, \kappa]$, we generate two pAPs of the form $a \cdot \vs^{(i)} < \theta$ with $a \in \{-1, 1\}$, yielding a total of $2\kappa$ pAPs. Note that the parameter $\theta$ has not yet been determined. We then use the pAPs to compose logic trees in five temporal forms:
\begin{align}
\bigcirc \mu,\, \lozenge \mu,\, \square \mu,\, \mu_1\, \mathcal{U}\, \mu_{2},\, \text{and }\, \mu_1 \mathcal{R}\, \mu_2,
\end{align}
where $\mu, \mu_1,$ and $\mu_2$ denote pAPs. This results in $6\kappa$ unary formulas, applying three unary operators to $2\kappa$ pAPs, and $8\kappa^2$ binary formulas, applying two binary operators to $4\kappa^2$ pAP pairs. We finally obtain a total of $6\kappa+8\kappa^2 $ basis logic trees. In this work, to reduce the construction overhead, we subsample basis logic trees by selecting a subset of state dimensions.

We then construct $N_\text{R}$ random logic trees by modifying the basis trees sampled to enhance the diversity of the initial population. For each, we begin by randomly selecting a basis tree and iteratively injecting a randomly chosen operator as the parent of a randomly selected node. This process continues until the tree exceeds a predefined depth $d_R$, or terminates early with probability $p_R$. In this work, we set $d_R=3$ and $p_R=0.1$. Finally, the size of the population becomes $N=N_\text{B}+N_\text{R}$. Note that, when $k>1$, the initial population includes the previously discovered $k-1$ TLTL constraints, replacing $k-1$ random trees to guide subsequent mining. 

\noindent\textbf{2) Parent selection}: This step selects $N_p$ parent logic trees based on the highest regularized fitness scores within the current population $\Phi=\{\phi_1, \ldots, \phi_N \}$ to generate offspring for the next generation. Let $\mathcal{V}(\phi_i)$ denote the node set of each tree $\phi_i$. The regularized fitness score $\mathcal{F}_{k}^\Phi(\phi)$ is:
\begin{align}
\mathcal{F}_{k}^\Phi(\phi)=\mathcal{F}_k(\phi)-\zeta
 \left(\frac{\sum_{i=1}^{|\Phi|}{\mathcal{F}_k(\phi_i)}}{|\Phi|} \right)
 \left(\, |\mathcal{V}(\phi)| -2\, \right)^{2},
\end{align}
where $\zeta$ is a hyperparameter and the $-2$ term accounts for the minimum number of nodes in a tree. To evaluate $\mathcal{F}_{k}^\Phi(\phi)$, we first compute the fitness score $\mathcal{F}_k$ by optimizing the parameters $\vtheta$ of $\phi$ via dual annealing~\cite{liu2024interpretable}. We then regularize the score by penalizing large tree sizes, encouraging the discovery of simpler constraints in this ill-posed mining problem~\cite{nenzi2018robust}.

\begin{figure}
  \centerline{\includegraphics[width=0.9\columnwidth]{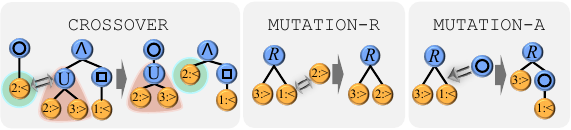}}
  \captionsetup{skip=-4.0pt}
  \caption{The illustration of the operations in the offspring generation. For each pTLTL tree in the figure, the blue nodes represent logical operators, and the orange nodes represent pAP nodes. The labels of the pAP node indicate the semantics of pAP; for example, the label '2:<' refers to $\vs^{(2)} < \theta$ for an undetermined parameter $\theta$.}
  \label{fig_offspring}
  \vspace{-20pt}
\end{figure}

\noindent\textbf{3) Offspring generation}: 
We iteratively generate $N$ offspring from the current population $\Phi = \{ \phi_1, \dots, \phi_{N_p} \}$ applying three genetic operators: \texttt{CROSSOVER}, \texttt{MUTATION-R}, and \texttt{MUTATION-A} (see Fig.~\ref{fig_offspring}). For each iteration, we first randomly select two parent trees, $\phi_i$ and $\phi_j$, using tournament selection. We then randomly apply a genetic operator to generate a new offspring in the following manner: 
\begin{itemize}[leftmargin=*]
\item \texttt{CROSSOVER}: we modify the selected trees, $\phi_i$ and $\phi_j$, by exchanging the randomly selected subtrees with each other. 
\item \texttt{MUTATION-R}: we randomly select a node in $\phi_i$ and either replace it with a same-type node\textemdash pAP or logic operator\textemdash or remove it along with its descendants.
\item \texttt{MUTATION-A}: we insert a randomly selected logic operator as the parent of a randomly chosen node in the tree $\phi_1$ preserving semantics.
(e.g., $\mu_1 \,\mathcal{U}\, \mu_2$ into $\mu_1 \,\mathcal{U}\, \square\mu_2$.)
\end{itemize}

Our method alternates parent selection and offspring generation, avoiding duplicates and enforcing temporal consistency by making every non-root node to have an ancestor with a temporal operator. For example, we disallow $\mu_1 \land \square \mu_2$, since $\mu_1$ lacks a temporal operator ancestor. We simplify logic trees using rules such as $\square\square=\square$ to encourage logical simplicity. Mining stops when the current logic accepts all expert demonstrations and rejects all other trajectories. In practice, we limit the number of mining steps to $4$ for the \textit{wiping} task and $5$ for the other tasks.

\vspace{-10pt}
\subsection{Policy player: logic-constrained reinforcement learning}
\label{sec:logic_CRL}
We introduce logic-constrained RL (Logic-CRL), which trains a policy to satisfy the TLTL constraints discovered by GA-TL-Mining. A main challenge is that TLTL constraints are non-Markovian; Their truths depend on entire state sequences rather than individual states. To resolve the mismatch, we embed each constraint into a product CMDP (PCMDP) by augmenting states with the TLTL constraint-based automaton states, thereby restoring the Markov property. Another challenge is the non-differentiable nature of logic constraints. To figure it out, we adopt Lagrangian CRL~\cite{tessler2018reward} with a novel state-wise constraint redistribution scheme, enabling conventional CRL frameworks to simplify the non-differentiable problem while still satisfying the non-Markovian TLTL constraint. We detail both processes in the following.  
%\vspace{-8pt}

\medskip
%%%%%%%%%%%%%%%%%%%%%%%%%%%%%%%%
%\noindent\textbf{Product Constrained Markov Decision Process}:
\noindent\textbf{PCMDP}:
We construct a PCMDP by synchronously combining the CMDP's transition system with a deterministic finite automata (DFA) converted from a TLTL constraint $\phi$. The DFA is the tuple $(\mathcal{Q}_\phi, \Sigma_\phi, \delta_\phi, q_{0,\phi}, F_\phi)$, where $\mathcal{Q}_\phi$ is the set of states, $\Sigma_\phi=2^{\textbf{AP}}$ is the power set of APs, $\delta_\phi: \mathcal{Q}_\phi \times \Sigma_\phi \rightarrow  \mathcal{Q}_\phi$ is the deterministic transition function, $q_{0,\phi} \in \mathcal{Q}_\phi$ is the initial state, and $F_\phi \subset \mathcal{Q}_\phi$ is the set of accepting states. 

The resulting PCMDP is the tuple, $\textstyle (\mathcal{S}_\phi, \mathcal{A}, \mathcal{L}_\phi, P_\phi, R, C_\phi)$, where $\mathcal{S}_\phi=\mathcal{S} \times \mathcal{Q}_\phi$ is the augmented state space, $\mathcal{L}_\phi:\mathcal{S}\rightarrow\Sigma_\phi$ is a labeling function assigning APs to states, and $P_\phi: \mathcal{S}_\phi \times \mathcal{A} \times \mathcal{S}_\phi \rightarrow \{0,1\}$ is the transition function defined as the product of the DFA's transitions and the CMDP's transition probabilities. The transition function deterministically advances the DFA state, starting from $q_{0,\phi}$, by evaluating the label $\mathcal{L}_\phi(\vs_{t+1})$ assigned to each next state $\vs_{t+1}$. Note that, in this work, we consider PCMDPs with optimal finite-horizon policies that prevent non-Markovian reward-seeking behaviors\textemdash such as staying in high-reward states and transitioning to acceptance states only at the last step\textemdash by assigning high rewards exclusively to acceptance states.

Finally, the PCMDP formulation synthesizes a policy that maximizes cumulative rewards subject to the TLTL constraint $\phi$. However, optimizing policies with TLTL constraints presents challenges due to their sparse nature and the delayed evaluation at the end of episodes. 

\noindent\textbf{Lagrangian CRL}: We introduce a Lagrangian CRL to handle sparse and temporally delayed TLTL constraint evaluation in policy learning. The sparse nature comes from Boolean semantics of TLTLs. To figure it out, we incorporate quantitative semantics with robustness function $\rho(\xi, \phi)$ to define a bounded but dense constraint-cost function $C_\phi' \in [0, 1]$, updating the discrete function $C_\phi$ in Eq.~\eqref{eq:icl_crl_obj}, as:
\begin{equation}
\label{eq:dense_constraint}
C_\phi'(\xi) = \alpha C_\phi(\xi) + \frac{1-\alpha}{\Upsilon} \mathrm{CLIP}[-\rho(\xi, \phi), 0, \Upsilon],
\end{equation}
where $\alpha\in[0,1]$ is a mixing weight, $\Upsilon\in\mathbb{R}^{+}$ limits the impact of large constraint violations, and $\mathrm{CLIP}$ truncates the negative robustness value within the range $[0, \Upsilon]$.

To address the delayed evaluation, we introduce a redistribution scheme that spreads out the trajectory-level constraint cost $C'_\phi(\xi)$ uniformly across all state-action pairs $(\vs_t,\va_t) \in \xi$, analogous to the uniform reward redistribution in IRCR~\cite{gangwani2020learning}. The resulting stepwise constraint cost $C'_\phi(\vs_t, \vq_t, \va_t)$ is in the augmented state space $(\vs_t, \vq_t) \in \mathcal{S}_\phi$ as 
\begin{equation} 
C'_\phi(\vs_t,\vq_t,\va_t) = \mathbb{E}_{\xi \sim \mathcal{D}} [ C'_\phi(\xi) | ((\vs_t,\vq_t),\va_t) \in \xi ],
\end{equation}
where $\mathcal{D}$ is the replay buffer containing previously collected trajectories in the augmented state space. This trajectory-space smoothing converts the trajectory constraint in Eq.~\eqref{eq:icl_crl_obj} into a state-action constraint, allowing its use in Lagrangian CRL. 

We formulate the Lagrangian dual of Eq.~\eqref{eq:icl_crl_obj} to solve for the constrained-optimal $\pi_k$ in the $k$-th Logic-CRL process:
\begin{align} 
\label{eq: smoothed lag CRL}
\mathop{\arg\min}_{\lambda \geq 0} \max_\pi \, \mathop{\mathbb{E}}_{\xi \sim \pi}  \quad\sum_{\mathclap{\substack{\vs_t,\va_t\sim\xi\\ \vq_{t} = \delta_\phi(\vq_{t-1},\mathcal{L_\phi}(\vs_t))} }} \; \left( 
R(\vs_t,\va_t) - \lambda \left(C'_{\phi}(\vs_t,\vq_t,\va_t)-d \right)\right),
\raisetag{1.4em}
\end{align}
where $\lambda$ is a learnable Lagrangian multiplier and $d$ is a constraint-cost threshold. We define a large threshold, $d=\frac{\varepsilon\alpha}{N_\xi}$, for stable optimization while enforcing the satisfaction of the binary logic constraint $C_\phi$, where $\varepsilon<1$ is a positive constant and $N_\xi$ is the number of trajectories sampled from the current policy $\pi$. To solve Eq.~\eqref{eq: smoothed lag CRL}, we adopt SAC-Lag~\cite{ha2021learning}\textemdash a soft actor-critic (SAC)~\cite{haarnoja2018soft} version of a Lagrangian CRL method.

\begin{figure*}
\centering
  \includegraphics[width=0.9\textwidth]{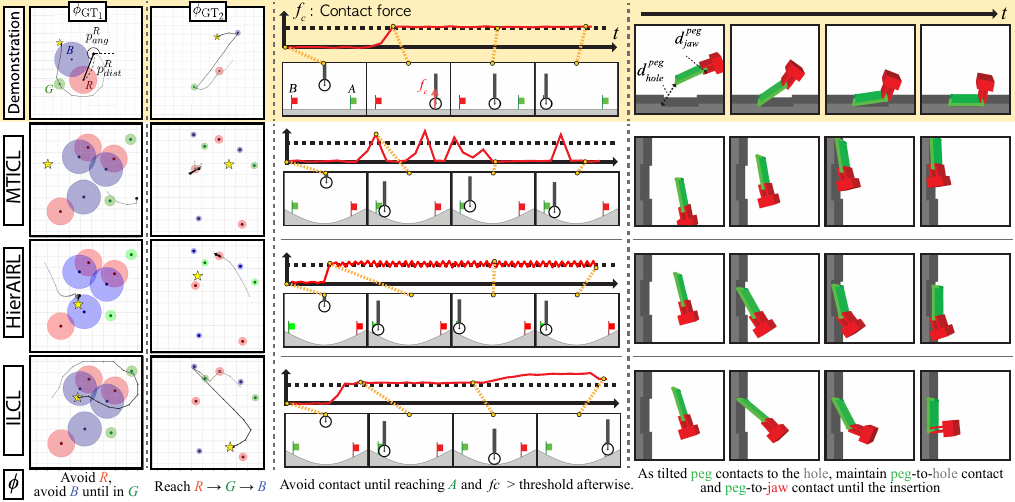}  
  \setlength{\abovecaptionskip}{0pt}
  \caption{Comparison of temporal constraint learning and transfer performance in four simulated tasks. In each task, ILCL, MTICL, and HierAIRL first learn temporal constraints from demonstrations (top), satisfying the ground-truth constraint $\phi$ (bottom). In novel environments, we train policies with the learned constraints or rewards to reproduce demonstration-like constrained behaviors. \textbf{Left}: in the \textit{navigation} tasks, yellow shapes and black traces represent goals and trajectories, respectively. \textbf{Middle}: in the \textit{wiping} task, green and red flags indicate start and goal locations, respectively. Red curve represents the observed contact force, and black-dot line represent the contact threshold from the demonstration. \textbf{Right}: in the \textit{peg-in-shallow-hole} task, red, green, and gray objects represent a gripper, a peg, and a hole, respectively. 
  }
  \label{fig_setup}
    \vspace{-15pt}
\end{figure*}

Further, to enhance off-policy learning in Eq.~\eqref{eq: smoothed lag CRL}, we initialize the replay buffer $\mathcal{D}$ with the expert trajectories $\Xi_E$ and the previously sampled trajectories $\Xi_{1:k-1}$ that already satisfy the given constraint. After obtaining $\pi_{k}$, we sample zero-violation trajectories $\Xi_{k}$ and update $\Xi_{1:k}= \Xi_{1:k-1} \cup \Xi_{k}$, which serves as input to the next GA-TL-Mining step.

Lastly, after $N$ alternations of GA-TL-Mining and Logic-CRL, ILCL returns the best constraint. The two-player game progressively finds constraints toward the ground truth when the constraint space is convex, and the constraint player's loss is linear and regularized~\cite{kim2024learning}. Although the logic constraint space is non-convex, we regularize the constraint player with tree size penalty as Eq.~\eqref{eq: GA-fitnessbase} to stabilize learning. We further choose the constraint $C_{\phi,i}$ that yields the most expert-like behavior from the obtained constraint set $\{ C_{\phi,0}, \dots, C_{\phi,N}\}$. For each candidate constraint, we evaluate its associated policy $\pi_{i+1}$ on the expert trajectories $\Xi_E$ and select the policy that minimizes the error to the expert actions,
\begin{equation}
    \pi_{i+1}=
    \underset{\pi_{i+1}\in[\pi_1,\ldots,\pi_{N+1}]}{\text{argmin}}
    \mathbb{E}_{\substack{(s_E,a_E)\sim\Xi_E\\ a\sim \pi_{i+1}(\cdot\mid s_E)}} \,\|a_E-a\|_2,
\end{equation}
and report the corresponding $C_{\phi,i}$.

\vspace{-8pt}
\section{Experimental Setup}
We conduct simulated benchmarks and demonstrate real-world transfer of the learned constraints.
\vspace{-12pt}

\subsection{Benchmark Setups}
\noindent\textbf{1) \textit{Navigation}}: This task requires a point agent at $\mathbf{x}^{Agt}\in\mathbb{R}^2$ navigating to a goal position $\mathbf{x}^{Goal}\in\mathbb{R}^2$ while satisfying temporal constraints related to colored regions randomly placed in the workspace (see Fig.~\ref{fig_setup} Left). Let $\mathbf{p}^{R}$, $\mathbf{p}^{G}$, and $\mathbf{p}^{B}$ denote the polar coordinates $(p_{dist}, p_{ang}) \in\mathbb{R}^2$ of the nearest red, green, and blue regions, respectively, measured relative to the agent's coordinate. For example, $\vp^{R}=(p^{R}_{dist}, p^{R}_{ang})$. We model this environment as two CMDPs, each subject to a distinct constraint, $\phi_{\text{GT1}}$ or $\phi_{\text{GT2}}$:
\begin{itemize}[leftmargin=*]
\item $\mathcal{S}=\{\vs \mid \vs=(\vx^{Agt}, \vx^{Goal}, \vp^{R}, \vp^{G}, \vp^{B})\in\mathbb{R}^{10}\}$.~
\item$\mathcal{A} = \{\va \,|\, \va = \Delta \vx^{Agt}\in\mathbb{R}^2\}$, the space of velocity commands.
\item $R = 1 - \tanh(5\cdot\Vert\vx^{Agt} - \vx^{Goal} \Vert_2)$
\item $C=C_{\phi_{\text{GT1}}}$ or $C_{\phi_{\text{GT2}}}$; $\phi_{\text{GT1}}$ requires the agent to always avoid red and avoid blue until reaching green. $\phi_{\text{GT2}}$ requires the agent to visit red, green, and blue in that order. 
\end{itemize}
We train each method on $20$ demonstrations ($T=25$), each constrained by $\phi_{\text{GT1}}$ and $\phi_{\text{GT2}}$ (see Table~\ref{table:best-small}). Then, we evaluate the learned constraints on test environments with varying number, size, and placement of regions. Each test uses a $1.5$ times larger workspace and a time horizon of $T=50$.

\noindent\textbf{2) \textit{Wiping}}: This task requires a roller agent at $\vx^{Agt}\in\mathbb{R}^2$ wiping the ground surface between the start point $A$ (green) and the goal point $B$ (red), as illustrated in Fig.~\ref{fig_setup} Middle. Let $f_c$ be the contact force between the roller and the surface. $^{A}\vx^{Agt}$ and $^{B}\vx^{Agt}$ denote the positions of the agent relative to the start and goal points, respectively. For example, $^{A}\vx^{Agt}=(^{A}x^{Agt}, ^{A}y^{Agt})$. We model this environment as a CMDP:
\begin{itemize}[leftmargin=*]
\item $\mathcal{S}=\{\vs \mid \vs=(^{A}\vx^{Agt}, ^{B}\vx^{Agt}, \dot{\vx}^{Agt}, w^{Agt}, d_g, f_g)\in\mathbb{R}^{9}\}$, where $\dot{\vx}^{Agt}$ and $w^{Agt}$ are the agent's linear and angular velocities, respectively. $d_g$ and $f_g$ are agent's distance and contact force from the ground surface, respectively. 
\item$\mathcal{A} = \{\va \,|\, \va = (f_x^{Agt}, f_y^{Agt})\in\mathbb{R}^2\}$; A space of force commands applied to the agent.
\item $R = R_{goal}+R_{contact}+R_{penalty}$; $R_{goal} = \ind( |^{B}x^{Agt}| < 0.1) - |^{B}x^{Agt}| \cdot \ind(|^{B}x^{Agt}| > 0.1)$ is a goal reward, $R_{contact}=2\cdot\ind(f_{c}>0.05)$ is a contact reward to encourage wiping contact, and $R_{penalty} = -0.2\cdot f_{c}^2$ is to penalize large contact.
\item $C=C_{\phi_{\text{GT}}}$ where the agent requires to keep sufficient contact force while the roller is between points $A$ and $B$, while avoiding contact otherwise.
\end{itemize}
We train each method on $100$ demonstrations ($T=50$), with randomized start, goal, and agent positions under constraint $\phi_{\text{GT}}$, and evaluate on test environments with sinusoidal surfaces of varying frequencies.

\noindent\textbf{3) \textit{Peg-in-shallow-hole}}: A gripper agent requires to insert a thin peg into a shallow-angled hole while maintaining side contact (see Fig.~\ref{fig_setup} Right). Let $\Sigma_{h}$ be the 2D hole frame tilted by angle $\theta^h$. The gripper at $^{\Sigma_{h}}\vx^{grip}$ is $(^{\Sigma_{h}}x^{grip}, ^{\Sigma_{h}}y^{grip}, ^{\Sigma_{h}}\theta^{grip})\in\mathbb{R}^3$, where $\theta^{grip}$ is the gripper tilt in the hole frame. The peg is at $^{\Sigma_{h}}\vx^{peg}$. We also define two features $d_{hole}^{peg}$ and $d_{jaw}^{peg}$ as the peg edge-to-hole and peg-edge-to-support-jaw distances, respectively. Then, a CMDP is 
\begin{itemize}[leftmargin=*]
\item $\mathcal{S}=\{\vs \mid \vs=( ^{\Sigma_{h}}\vx^{grip}, ^{\Sigma_{h}}\dot{\vx}^{grip}, ^{\Sigma_{h}}\vx^{peg}, ^{\Sigma_{h}}\dot{\vx}^{peg}, ^{\Sigma_{h}}\bar{\vx}^{grip}, \\ d^{jaw}, \dot{d}^{jaw},   \bar{d}^{jaw} , d_{hole}^{peg}, d_{jaw}^{peg}, \cos\theta^h, \sin\theta^h )\in\mathbb{R}^{22}\}$, where $^{\Sigma_{h}}\bar{\vx}^{B}$ is the desired pose for the next time step. $d^{jaw}$, $\dot{d}^{jaw}$ and $\bar{d}^{jaw}$ are the position, velocity, and desired position of the jaws, respectively. 

\item$\mathcal{A} = \{\va \,|\, \va = (\Delta^{\Sigma_{h}}\bar{\vx}^{grip}, \Delta\bar{d}^{finger} )\in\mathbb{R}^4\}$; A space of discrete-time velocity commands for the gripper and fingers. 
\item $R=(1-\tanh(5|^{\Sigma_h}y^{peg} - ^{\Sigma_h}y^{peg*}|)) + 5 \cdot \ind(^{\Sigma_h}y^{peg} - ^{\Sigma_h}y^{peg*})$, where $^{\Sigma_h}y^{peg}$ and $^{\Sigma_h}y^{peg*}$ are the current and desired vertical position of the peg in terms of the hole frame $\Sigma_h$. 
\item $C=C_{\phi_{\text{GT}}}$ where the agent requires maintaining peg-to-hole contact once tilting the peg above a certain angle, while keeping the peg-to-jaw contact until full insertion.
\end{itemize}
We train each method on $100$ demonstrations ($T=30$), with randomized start, goal, agent poses as well as randomly tilted holes ($\theta^h\in[-10^\circ, 10^\circ]$), under the constraint $\phi_{\text{GT}}$. Then, we evaluate on test environments with tilt angles $\theta_h\in\{-90^\circ. -45^\circ, 45^\circ, 90^\circ\}$ increasing the time horizon into $T=60$. 

\vspace{-10pt}
\subsection{Quantitative Evaluation Study with Baselines}
We evaluate each methodology in both training (seen) and testing (unseen) environments. For training, we estimate a constraint and the corresponding constrained policy for each random seed. From each policy, we generate $10^3$ trajectories by randomizing initial conditions. In total, we produce $10^4$, $5\cdot 10^3$, and $5\cdot 10^3$ trajectories using $10$, $5$, and $5$ random seeds, respectively, for the three tasks presented. 

For testing, we transfer the learned constraints (i.e., $10$, $5$, and $5$ constraints per task) to novel, randomly generated environments ($10$, $4$, and $4$ environments per task). We train one constrained policy per environment and sample $10^3$ trajectories with varied starts or goals, yielding $10^5$, $2\cdot 10^4$, and $2\cdot 10^4$ trajectories. We then assess performance using the followings:
\begin{itemize}[leftmargin=*]
\item \textbf{VR}: the violation rate $[\%]$ for a ground-truth constraint $\phi_{\text{GT}}$ where any timestep violation marks the episode as violated.
\item \textbf{REW}: the average of episode rewards from trajectories.
\item \textbf{TR}: the average of truncated negative robustness values, where each value comes from  $\max (-\rho(\xi, \phi_{\text{GT}}),0)$,
\end{itemize}
We compare ILCL against ICL and inverse reinforcement learning (IRL). The ICL methods include numeric state-action (i.e., Markovian) constraint learning approaches.
\begin{itemize}[leftmargin=*]
    \item \textbf{ICRL}~\cite{malik2021inverse}: An inverse CRL (ICRL) method with maximum likelihood constraint inference.
    \item \textbf{TCL}~\cite{jang2023inverse}: A transferable constraint learning (TCL) method by reward decomposition.
    \item \textbf{MTICL}~\cite{kim2024learning}: A multi-task ICL (MTICL) method with a two-player zero-sum game scheme (i.e., basis for ILCL).
\end{itemize}
The IRL methods, particularly hierarchical IRL (HIRL), generate policies with multiple level temporal abstraction: 
\begin{itemize}[leftmargin=*]
    \item \textbf{Option-GAIL} \cite{jing2021adversarial-optiongail}: An HIRL extension of generative adversarial imitation learning~(GAIL).
    \item \textbf{Hier-AIRL}~\cite{chen2023option-hierairl}: An HIRL extension of adversarial inverse reinforcement learning (AIRL).
\end{itemize}

\begin{figure*}[!t]
\centering
 \includegraphics[width=0.95\textwidth]{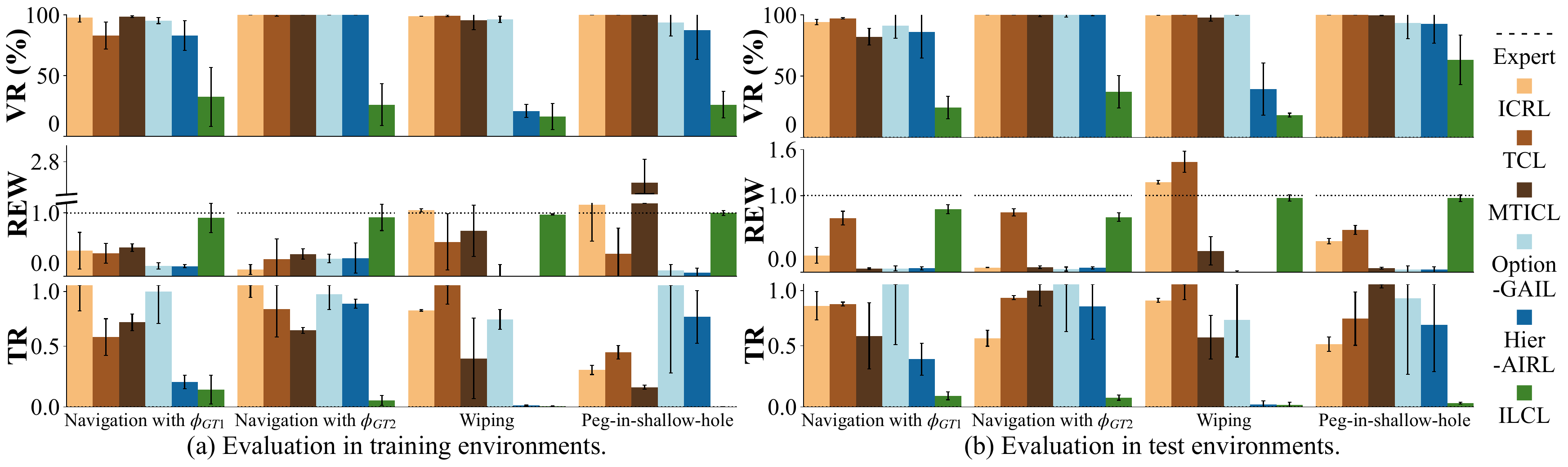}
 \captionsetup{skip=-4pt}
  \caption{Comparison of the proposed ILCL and baseline methods in training and test environments. `Expert' denotes the result of expert demonstrations without constraint violations. We normalize all REW values to the `Expert' score, resulting in a range of $[0,1]$.}
  \label{fig: combined bars}
  \vspace{-15pt}
\end{figure*}

For evaluations, all methods receive the same features for the ground-truth constraints. IRL baselines additionally get reward-related features to learn their rewards.

\vspace{-10pt}
\subsection{Qualitative Transfer Study in Real World}
We perform qualitative transfer studies in real-world \textit{peg-in-shallow-hole} environments, as shown in Fig.~\ref{fig_main}. The environment consists of a 7-DoF Franka Emika Panda arm, an external RGB-D camera, a peg, and a tilt of shallow hole. After learning the stable-contact constraint through simulations, we enable the robot to reproduce demonstration-like constrained behaviors via Logic-CRL in novel environments. 

To show the transferability, we vary the hole's slope within $\{-25^\circ, 60^\circ, 90^\circ\}$. To detect slope and peg states, we use AprilTags on the gripper, hole, and peg. The learned policy applies discrete-time velocity commands, sending desired poses to a Cartesian PD controller at \SI{1}{\kilo\hertz} and RL controller at \SI{5}{\hertz}.

\vspace{-8pt}
\section{Evaluation Results}
\subsection{Quantitative Analysis through Simulations}
% Logic Learning
We evaluate the constraint learning performance of ILCL against baselines across four benchmark environments with distinct ground-truth logic constraints. As shown in Fig.~\ref{fig: combined bars}~(a), ILCL consistently outperforms baselines in learning constraints, achieving the lowest VRs in all training environments while obtaining expert-like REWs. We normalize all REW values to the expert demonstrations' scores, ranging from $[0,1]$. Notably, in the \textit{navigation} scenario with $\phi_{\text{GT1}}$, ILCL's maximum violation rate of $32.5\%$ is $50.4\%$ lower than Hier-AIRL's, the best performing baseline. In contrast, ICL baselines exhibit significantly higher VRs due to their inability to learn temporal structure. Although IRL baselines learn the hierarchical structure in demonstrations, the approaches often fail to capture either the task objective or the embedded constraint. For example, Hier-AIRL in the \textit{wiping} task achieves a VR of $21\%$, lower than other baselines, but still yields low REWs due to a deficient task reward signal. In contrast, REWs exceeding $1.0$ indicate less- or un-constrained goal-seeking behaviors due to failures in constraint learning or CRL. Note that the binary nature of VRs classifies even minor, single timestep violations as full binary violations. A continuous measure is necessary to provide a more detailed assessment of constraint quality.
\newcolumntype{A}{>{\centering\arraybackslash}m{0.9cm}<{}}
\newcolumntype{L}{>{\centering\arraybackslash}m{1.0cm}<{}}
\newcolumntype{T}{>{\centering\arraybackslash}m{1.3cm}<{}}

\setlength{\textfloatsep}{0pt}{
\begin{table}[t]
\caption{Ground-truth (GT) and the lowest VR constraints from ILCL.
Reported statistics share the same scales with Fig.~\ref{fig: combined bars}.
}
\vspace{1pt}
\scriptsize
\setlength{\tabcolsep}{1pt}
\centering
\begin{tabular}{@{}>{\centering\arraybackslash}p{0.8cm} c c c c c@{}}
\toprule[1pt]
Env. & Algo. & \multicolumn{1}{c}{TLTL Const.} & VR (\%) & REW & TR \\
\midrule
\multirow{3}{*}{\makecell[c]{\textit{Nav.}\\$\phi_{\text{GT1}}$}} 
& \scriptsize GT
& \scalebox{0.9}{$\square p^{R}_{\text{dist}} > 0.2 \land (p^{B}_{\text{dist}} > 0.25 \, \mathcal{U} \, p^{G}_{\text{dist}} < 0.08)$}
& 0 & 1 & 0 \\
\cmidrule(lr){2-6}
& \scriptsize ILCL 
& \scalebox{0.9}{$ (\square p^R_{\text{dist}} > 0.205 \land p^B_{\text{dist}} > 0.267) \mathcal{U} p^G_{\text{dist}} < 0.081$} 
& 1.3 & 0.83 & 2.6E-5 \\
\midrule
\multirow{3}{*}{\makecell[c]{\textit{Nav.}\\$\phi_{\text{GT2}}$}} 
& \scriptsize GT
& \scalebox{0.9}{$\lozenge (p^{R}_{\text{dist}} < 0.06 \land \lozenge(p^{G}_{\text{dist}} < 0.05 \land \lozenge(p^{B}_{\text{dist}} < 0.04)))$}
& 0 & 1 & 0 \\
\cmidrule(lr){2-6}
& \scriptsize ILCL 
& \scalebox{0.9}{$(\lozenge{p^B_{\text{dist}} < 0.041}\, \mathcal{U}\, p^G_{\text{dist}} < 0.049)\, \mathcal{U}\,p^R_{\text{dist}} < 0.061$} 
& 18.5 & 0.69 & 3.5E-2 \\
\midrule
\multirow{5}{*}{\makecell[c]{\textit{Wiping}}}
& \multirow{2}{*}{\scriptsize GT}
& \scalebox{0.9}{$f_{\text{c}}<0.05 \mathcal{U} (\square f_{\text{c}}>1.6 \, \mathcal{R}\,$} & \multirow{2}{*}{0} & \multirow{2}{*}{1} & \multirow{2}{*}{0} \\
& 
& \scalebox{0.9}{$(^Ax^{Agt} > -0.1 \land {^A}x^{Agt} < 0.1))$} &&& \\
\cmidrule(lr){2-6}
& \multirow{2}{*}{\scriptsize ILCL}
& \scalebox{0.9}{$\square f_c > 1.661\, \mathcal{R}\, (f_c < 0.062 \,\mathcal{U}$} & \multirow{2}{*}{\underline{\textbf{0}}} & \multirow{2}{*}{0.97} & \multirow{2}{*}{\underline{\textbf{0}}} \\
& 
& \scalebox{0.9}{$({}^A x^{Agt} < 0.099 \land {}^A x^{Agt} > -0.076))$} &&& \\
\midrule
\multirow{6}{*}{\makecell[c]{\textit{Peg-in-}\\\textit{shallow}\\\textit{-hole}}}
& \multirow{2}{*}{\scriptsize GT}
& \scalebox{0.9}{$\lozenge(^{\Sigma_h}\theta^{peg} >34.88^\circ\land (^{\Sigma_h}\theta^{peg} < 4.217^\circ$} & \multirow{2}{*}{0} & \multirow{2}{*}{1} & \multirow{2}{*}{0} \\
& 
& \scalebox{0.9}{$\mathcal{R}\, d_{jaw}^{peg} < 0.0053 )\land \square d_{hole}^{peg} < 0.0072)$} &&& \\
\cmidrule(lr){2-6}
& \multirow{2}{*}{\scriptsize ILCL} 
& \scalebox{0.9}{$\lozenge ({}^{\Sigma_h}\theta^{\text{peg}} > 34.68^\circ \land {}^{\Sigma_h}\theta^{\text{peg}} < 39.44^\circ \land ({}^{\Sigma_h}\theta^{\text{peg}} $} & \multirow{2}{*}{11.2} & \multirow{2}{*}{0.95} & \multirow{2}{*}{2.1E-4} \\
& 
& \scalebox{0.9}{$< 0.241^\circ\,\mathcal{R}\, d^{\text{peg}}_{\text{jaw}} < 0.0055) \land \square d^{\text{peg}}_{\text{hole}} < 0.0072)$} &&& \\
\bottomrule[1pt]
\end{tabular}
\label{table:best-small}
\end{table}}

Alternatively, to provide a more detailed assessment beyond the binary measure of violations, we investigate the degree of violations using the TR metric. ILCL consistently demonstrates the lowest mean TR values, significantly outperforming all baselines. These results underscore the complexity of temporal-constraint learning, even in seen environments, for conventional ICL or IRL algorithms. Although the ICRL and MTICL approaches yield high REWs in the last two tasks, these rewards are from unconstrained behaviors, resulting in significantly high violation rates, which are undesirable in robotic tasks. In addition, IRL shows the lowest REWs across most tasks, reflecting the difficulty of simultaneously optimizing both task rewards and constraints.

We then assess the transferability of the learned constraints to the novel environments as shown in Fig.~\ref{fig: combined bars}~(b). ILCL consistently achieves the lowest VR and TR values with high REWs, similar to its performance during training. In contrast, all baselines exhibit VRs exceeding $85.9\%$, with the sole exception of Hier-AIRL in the \textit{wiping} task, highlighting the significant challenges in transferring temporal constraints. Although Hier-AIRL yields a modest VR in \textit{wiping}, it frequently gets stuck at the start point as illustrated in Fig.~\ref{fig_setup} and fails to collect task rewards in test environments. Further, in the \textit{peg-in-shallow-hole} task, ILCL shows a relatively high VR of $63.3\%$ despite a significantly low TR value. This discrepancy arises not only from the binary nature of VR but also from the Lagrangian approximation used in Logic-CRL, which softens constraints to prioritize high-reward behaviors, increasing violations in novel settings.

Table~\ref{table:best-small} shows examples of ground-truth and estimated constraints as well as the exemplar performance of their corresponding constrained policies per test environment. The examples show that ILCL returns a constraint similar to the ground truth. Although the estimated constraints exhibit syntactic variations from the ground truth, they consistently achieve significantly low VRs, averaging $7.8\pm8.7\%$, and high REWs, averaging $0.86\pm0.13$, similar to ground-truth performance across all tasks. To ensure accurate comparison, we evaluate $1,000$ trajectories that do not violate the target constraints used in Logic-CRL, thereby reducing its potential negative impact on this analysis.

\begin{figure}[t]
    \centering
    \includegraphics[width=0.98\linewidth]{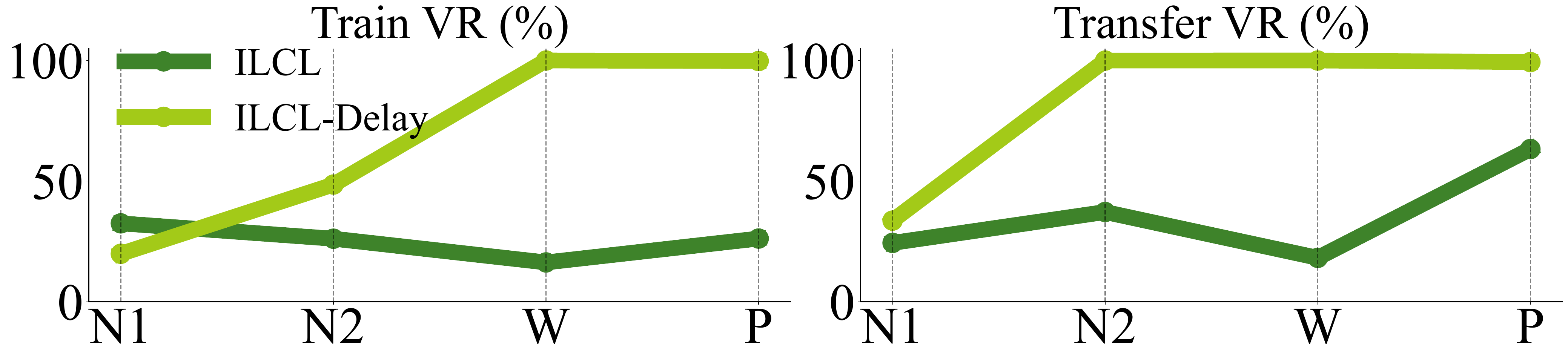}    
    \captionsetup{skip=-2pt}
    \caption{Comparison of mean VR between ILCL and ILCL-Delay in training (left) and test (right) environments. N1, N2, W, and P indicate the evaluation tasks in the same order as in Fig. 5.}
    \label{fig:ablation}
\end{figure}

Lastly, to quantify redistribution in Logic-CRL, we compare ILCL with an ablation that removes the redistribution, which we denote as ILCL-Delay. Fig.~\ref{fig:ablation} shows ILCL-Delay exhibits high VRs in sparse constraint signal tasks, such as wiping and peg-in-shallow-hole, and even reaches $100\%$ VR on the transfer experiment of navigation with $\phi_{GT2}$, highlighting the difficulty of learning new policies without redistribution. These demonstrate that constraint redistribution is crucial for handling non-Markovian temporal constraints.

\vspace{-8pt}
\subsection{Qualitative Demonstrations in Real World}
\begin{figure*}[t]
\centering
    \includegraphics[width=0.91\textwidth]{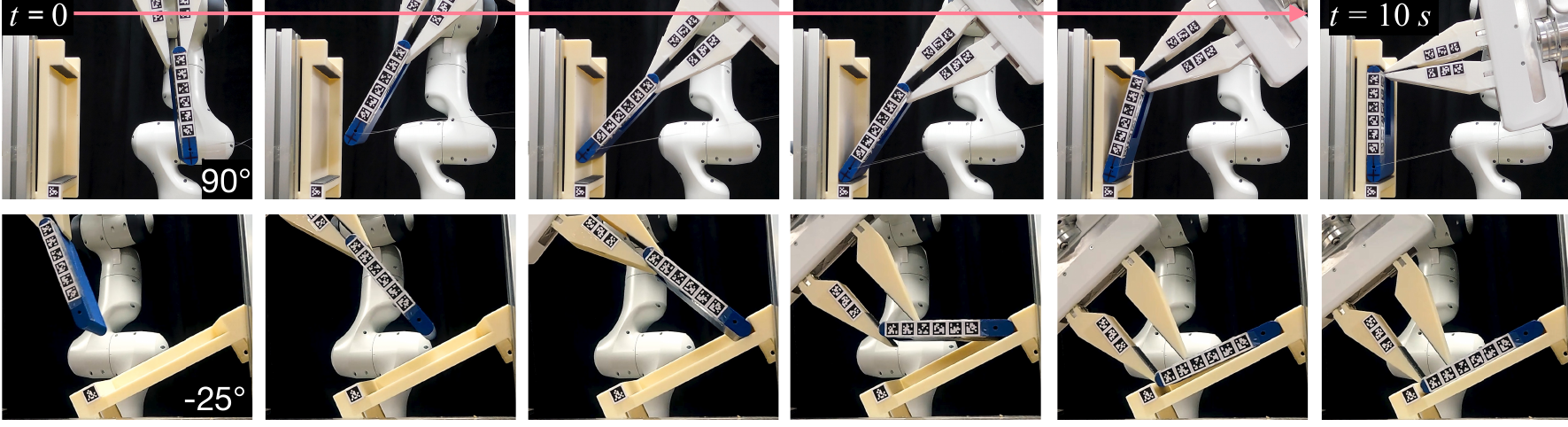}
  \caption{
  Demonstration of ILCL's constraint transfer capability in the \textit{peg-in-shallow-hole} environments. By creating real-world environments with holes tilted at $90^\circ$, $60^\circ$, and $-25^\circ$, we enable the Panda gripper to find a constrained policy using the learned constraints and reproduce a demonstration-like robust insertion with stable contact in each unseen environment. Demonstration videos are available in the supplementary materials including $60^\circ$ tilted hole insertion.}
  \label{fig_demonstrations}
  \vspace{-18pt}
\end{figure*}

% Real-world demonstration
Fig.~\ref{fig_demonstrations} demonstrates the transferability of ILCL in novel \textit{peg-in-shallow-hole} environments, where the Panda arm with a parallel jaw gripper successfully inserts a peg into unforeseen holes tilted at $90^\circ$, $60^\circ$, and $-25^\circ$. During the insertions, the gripper maintained stable contact between the hole and the peg. These demonstrations indicate that ILCL learns abstract logic constraints that are robustly transferable to real-world setups. For each experiment, we optimized the constrained policy for $4$ hours within $10,000$ steps of real-world interactions.

\vspace{-8pt}
\section{Conclusion}
\vspace{-4pt}
We introduced inverse logic-constraint learning (ILCL) that identifies free-form transferable TLTL constraint from demonstrations. Adopting a two-player zero-sum game, ILCL combines GA-TL-Mining\textemdash learning TLTL constraints from demonstrations without predefined templates\textemdash and Logic-CRL\textemdash optimizing policies under non-Markovian constraints. In quantitative and qualitative studies, ILCL outperforms state-of-the-art baselines, reproducing demonstration-like behaviors with fewer violations in novel environments, and generalizes to real-world peg-in-shallow-hole tasks. 
\vspace{-10pt}

%\section*{Acknowledgments}
%%%%%%%%%%%%%%%%%%%%%%%%%%%%%%%%%%%%%%%%%%%%%%%%%%%%%%%%%%%%%%%%%%%%%%%%%%%%%%%%
\bibliographystyle{IEEEtran}
\bibliography{references_new}
%%%%%%%%%%%%%%%%%%%%%%%%%%%%%%%%%%%%%%%%%%%%%%%%%%%%%%%%%%%%%%%%%%%%%%%%%%%%%%%%
\clearpage  
\end{document}